\documentclass[journal]{IEEEtran}

\usepackage[cmex10]{amsmath}
\usepackage{amsthm}
\usepackage{amssymb}
\usepackage{mathrsfs}
\usepackage{graphicx}
\usepackage{float}
\usepackage{array}
\usepackage{epstopdf}
\usepackage{multirow}
\usepackage[justification=centering]{caption}

\begin{document}

\title{\Large \textbf{IDENTIFYING MILD TRAUMATIC BRAIN INJURY PATIENTS FROM MR IMAGES USING BAG OF VISUAL WORDS}}

%{\dagger}, {^{**}} 
\author{Shervin~Minaee$^{1}$, Siyun Wang$^1$, Yao~Wang$^1$, Sohae Chung$^2$, Xiuyuan Wang$^2$, Els Fieremans$^2$, Steven Flanagan$^3$, Joseph Rath$^3$, Yvonne W. Lui$^2$,  \\
$^1$Electrical and Computer Engineering Department, New York University
\\ $^2$Department of Radiology, New York University
\\ $^3$Department of Rehabilitation Medicine, New York University}

\maketitle

\IEEEpeerreviewmaketitle

\vspace{-0.5cm}

%\section{\fontsize{16}{15}\selectfont purpose}
\begin{abstract}
Mild traumatic brain injury (mTBI) is a growing public health problem with an estimated incidence of one million people annually in US. 
Neurocognitive  tests are used to both assess the patient condition and to monitor the patient progress. 
This work aims to directly use MR images taken shortly after injury to detect whether a patient suffers from mTBI, by incorporating machine learning and computer vision techniques to learn features suitable discriminating between mTBI and normal patients.
We focus on 3 regions in brain, and extract multiple patches from them, and use bag-of-visual-word technique to represent each subject as a histogram of representative patterns derived from patches from all training subjects.
After extracting the features, we use greedy forward feature selection, to choose a subset of features which achieves highest accuracy.
We show through experimental studies that BoW features perform better than the simple mean value features which were used previously.
\end{abstract}

\section{Introduction}
Mild traumatic brain injury (mTBI) is a growing public health problem, which can cause loss of consciousness and/or confusion and disorientation.
In addition to civilian head trauma, we are now faced with on-going U.S. military-related brain injury as well as greater numbers of sport-related head injuries \cite{mtbi1}. 
The person with mTBI usually has cognitive problems such as headache, difficulty thinking, memory problems, attention deficits, mood swings and frustration.
Up to 20-30\% of patients with mTBI develop persistent symptoms months to years after the initial injury, referred to as post-concussive syndrome (PCS), resulting in substantial disability. 

Currently, several different definitions of mTBI exist (World Health Organization, American Congress of Rehabilitation Medicine [ACRM] \cite{acrm},  Centers for Disease Control and Prevention \cite{cdcp},
Department of Defense \cite{dod}, and Department of Veteran Affairs \cite{dva}). 
There is universal agreement that a unified, objective definition is needed. 
Furthermore, most identification schemes rely on Glasgow Coma Scale score, which was recently deemed insufficient for diagnosing  traumatic brain injury by the National Institute for Neurological Disorders and Stroke, which proposed that neuroimaging have a larger role in the classification scheme for mTBI. 
Recent work using MRI revealed that there are areas of subtle brain injury after mTBI; however, no single imaging metric has thus far been shown to be sufficient as an independent biomarker.

While diffusion MRI has been extremely promising in the study of mTBI, identifying patients with recent mTBI remains a challenge.
The literature is mixed with regard to localizing injury in these patients, however, gray matter such as the thalamus and white matter including the corpus callosum and frontal deep white matter have been repeatedly implicated as areas at high risk for injury.
In \cite{yuanyi}, Lui proposed a machine learning approach based on mean feature values of different metrics from MR images.
In \cite{vergara}, Vergara proposed an approach based on features derived from resting state functional network connectivity (rsFNC) and diffusion magnetic resonance imaging, followed by linear support vector machine.
While these works are also using a machine learning approach, but the feature used for them may not be the best set for this task.

The purpose of this study is to develop a machine learning framework to classify mTBI patients and controls using features derived from multi-shell diffusion MRI in the thalamus, frontal white matter and corpus callosum.
In the machine learning community, it is well known that using multiple features can improve classification performance compared with a single feature alone, and that the performance of classification algorithm mainly relies on the usefulness of feature set.
We have explored a new approach for feature extraction from MR images, where instead of the prior approach where the mean value of different metrics in various brain regions are used as feature, we use computer vision based techniques to learn a set of visual words from diffusion MR images of brain, using bag-of-visual-word (BoW) approach \cite{bow1}. 
We then use feature selection followed by a classification algorithm to identify mTBI patients.
We show that by using greedy forward feature selection, we are able to achieve higher accuracy over single best feature.
%Although this methodology works to some extent, it is not guaranteed to achieve the highest performance. 
Through experimental study, we show that these features result in much higher accuracy compared to the simple mean features. 
The preliminary results of this work are presented in \cite{afsnr}.
This approach provides a powerful scheme to learn a global representation by aggregating local information, and is especially useful for datasets with limited number of samples but high dimensional input data \cite{fr1}-\cite{fr2}.

The rest of the paper is organized as follows. 
Section II provides a description of the proposed framework. 
Section III provides a brief overview of bag of visual words approach. 
Section IV provides the experimental studies and performance analysis. 
And finally the paper is concluded in Section V.

\section{The Proposed Framework}
There have been some previous works on mTBI classification using various sets of features, from demographic (such as age and gender) and neurocognitive  to imaging related features.
Demographic features alone would not be sufficient to classify a person as mTBI, and it would be helpful to include all possible features in the feature pool and use feature selection to pick the best subset of features. 
Demographic and neurocognitive  features are easy to derive, but for imaging features, it is not clear what is the best way to derive them.
For demographic features, age and sex are used in this paper. 
And for neurocognitive features, we used Stroop, Symbol Digit Modalities Test (SDMT), California Verbal Learning Test (CVLT) and Fatigue Severity Scale (FSS).
In the past few years, we have developed specialized MR imaging protocols and related image features that are promising for distinguishing mTBI patients from controls \cite{lui1}-\cite{lui5}. 
Some of these metrics are summarized in Table I. 
\begin{table}[h]
\centering
\begin{tabular}{| m{1.5cm} | m{4.5cm} |}
\hline
MRI Metric & Metric Description  \\
\hline
AWF & Axonal Water Fraction  \\
\hline
DA & Diffusivity within Axons  \\
\hline
De-par & Diffusion parallel to the axonal tracts in the extra-axonal  \\
\hline
De-perp & Diffusion perpendicular to the axonal tracts in the extra-axonal  \\
\hline
FA & Fractional Anisotropy  \\
\hline
MD & Mean Diffusion  \\
\hline
AK & Axial Kurtosis  \\
\hline
MK & Mean Kurtosis  \\
\hline
RK & Radial Kurtosis  \\
\hline
\end{tabular}
\vspace{0.2cm}
\caption{MRI metrics description}
  \label{tab:1}
\end{table}

One way to derive features from MR images, is to calculate the mean value of (some of ) the above metrics in different regions such as: thalamus, prefrontal white-matter, corpus callosum (CC) Body, CC-Genu, and CC-Splenium (which are focused in our work).
But mean value may not be the best way to extract features from a specific region and metric.
In this work, we propose a new approach for learning features from MR images, based on bag-of-visual-words.
This approach is explained in Section III.
After extracting features, we use feature selection \cite{sel} to reduce the dimensionality of the feature.
We tried multiple greedy approaches for feature selection, and found out greedy forward feature selection performs best for this task.
Greedy forward feature selection selects the best features one at a time. Assuming $S_k$ denotes the best subset of features of size $k$, the $(k+1)$-th feature is selected as the one which results in the highest cross-validation accuracy rate along with the features already chosen (in $S_k$).
One can stop adding feature, either by setting a maximum size for the feature set, or when adding more features does not increase the accuracy rate.
The block diagram of the overall algorithm is shown in Figure 1.
As we can see the features from image are concatenated with selected demographic and neurocognitive features and used for classification.
After selecting the feature subset, a classification algorithm is used to classify the samples into patient and control. 
Different classifiers can be used for this purpose, such as support vector machine (SVM) \cite{svm}, logistic regression \cite{lr}, random forest \cite{rf}, and neural network \cite{nn}.
Based on our experimental studies, we found that most of these classifiers results in similar accuracy, but SVM achieves slightly higher accuracy. Therefore we performed most of our experimental studies using SVM.

\begin{figure}[h]
\begin{center}
    \includegraphics [scale=0.42] {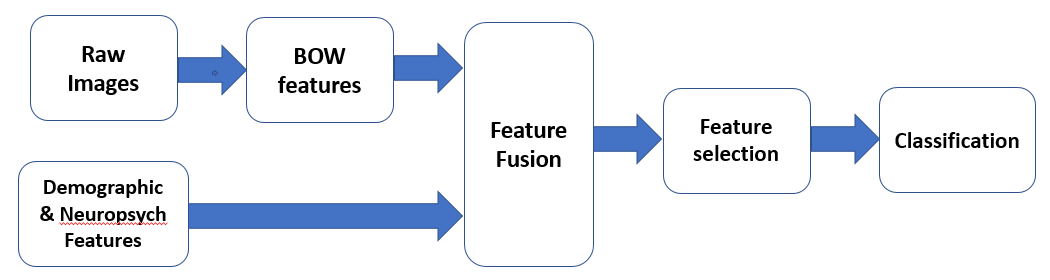}
\end{center}
\vspace{-0.2cm}
  \caption{Schematic of ``Bag of Words'' approach.}
\end{figure}

\iffalse
In the first approach, mean values of above 7 MRI metrics in five regions (thalamus, prefrontal white-matter, corpus callosum (CC) Body, CC-Genu, and CC-Splenium) were used in the feature representation along with two demographic features (age and sex) and four neurocognitive tests (Stroop, Symbol Digit Modalities Test (SDMT), California Verbal Learning Test (CVLT) and Fatigue Severity Scale (FSS)). 
A greedy forward feature selection approach was used to choose the best feature subset upon which a support vector machine achieves the highest cross-validation accuracy.
\fi

\section{region-specific bag-of-words representation for MR images}
Bag of visual words is a popular approach in computer vision \cite{bow1}, which is used for various applications \cite{bow2}-\cite{bow3}.
The idea of bag of visual words in computer vision is inspired by bag of word representation in text analysis, where a document is represented as a histogram of words from a dictionary, and these histograms are used to analyze the text documents \cite{bow_text}.
In the same way, one can represent an image (or video) as a histogram of visual words.
Since there is no intrinsic words defined for images, we need to first create the visual words. 
A popular approach is to extract a large number of patches from training images (either around key-points, or over a regular grid), and then use clustering algorithms, such as k-means \cite{kmeans} and mean-shift \cite{mshift}, to cluster these patches into $K$ clusters, and use their centroids as the visual patterns. 
Instead of raw pixel values in patches, one can also extract some image descriptor from each patch and learn the words from those features.
Then to derive the BoW representation for a new image, it is first divided into several patches, and then the histogram of those patches are found over the visual words (learned over training samples), and these histograms are used as the feature representation of the image (or video). 
Figure 2 denotes the schematic of the BoW algorithm for brain images.
\begin{figure}[h]
\begin{center}
%\captionsetup{justification=centering}
    \includegraphics [scale=0.35] {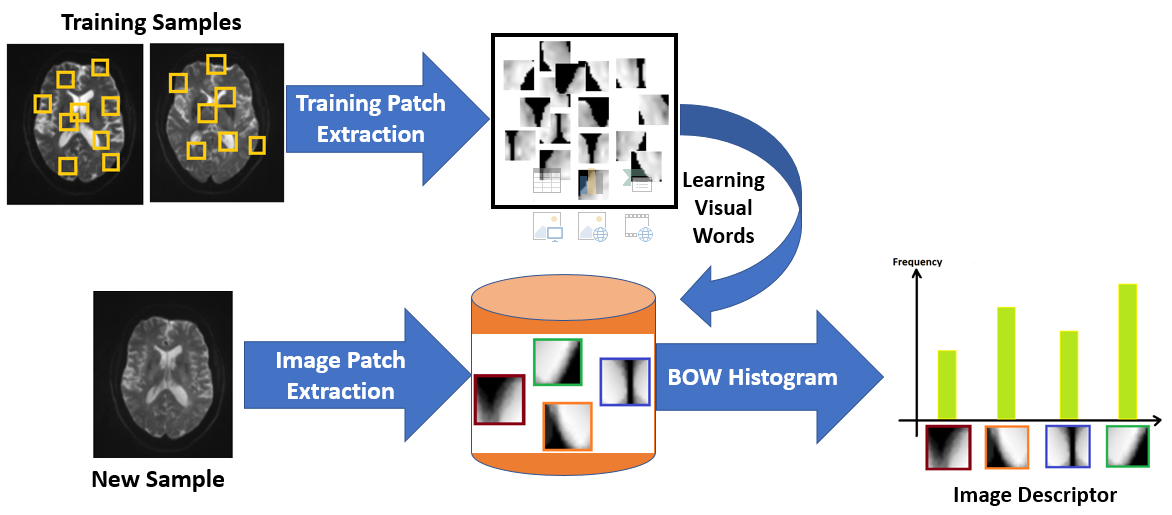}
\end{center}
\vspace{-0.2cm}
  \caption{Schematic of ``Bag of Words'' approach.}
\end{figure}

In our case, 16x16 patches are extracted from brain slices through the areas of interest and all the training patches in the mTBI patients and control subjects are separately clustered to learn the most representative visual patterns (called “visual” words). 
For our problem, we applied BoW approach on various imaging metrics (AWF, DA, De\_par, De\_perp, FA, MD, AK, MK, RK) for two brain regions, Corpus Callosum and Thalamus (for Thalamus only FA, AK, MK, RK, MD metrics are used), and learned a set of 20 visual words from patches of size 16x16 for each one of them  for  mTBI and control populations separately. 
K-means clustering is used to learn visual words in each case.
Then for each subject, we extract patches from the two brain regions, and for each patch find the closest one among all words in both dictionary. 
Finally we concatenate the BoW histograms for all metrics and two regions to derive the final visual representation (a 220-dimensional feature).

\section{experimental results}
We collected a set of 69 mTBI subjects between 18 and 64 years old, within 1 month of mTBI as defined by the American College of Rehabilitation Medicine (ACRM) criteria for head injury and 40 healthy age and sex-matched controls. 
%Imaging was performed on a 3.0 Tesla Siemens Trio (Erlangen, Germany) magnet including multi-shell diffusion MRI at 5 b-values (250, 1000, 1500, 2000, 2500 s/mm2) in a total of 136 directions using multiband 2 at isotropic 2.5mm image resolution.

To evaluate the model performance, we use a similar approach to 5-fold cross validation, where each time we take 20\% of the samples for validation, and the rest for training.
For the forward feature selection process, for each candidate new feature, the training samples are used to train the SVM model, and the performance is evaluated on validation samples.
To decrease the sampling bias, we repeat this approach 50 times, and take the average accuracy as the cross validation accuracy.
For SVM, we use radial basis function (RBF) kernel. 
The hyper-parameters of SVM model (kernel width gamma, and the mis-classification penalty weight, C) are tuned to achieve the highest cross validation accuracy for each candidate feature set.
It is worth to mention that, we normalize all features before feeding as the input to SVM, by making them zero-mean and unit-variance. 
%For histogram features, we skip this normalization (as they are already normalized).

For the baseline, we use a feature set that includes the mean value of metrics in different regions (as in \cite{yuanyi}), along with demographic and neurocognitive  features, the best single feature achieves a classification accuracy of 72\% (in cross-validation sense), using AWF in CC-Body. 
Also, the best feature subset chosen by the greedy feature selection has an accuracy of 80\% with 8 features (De\_par in thalamus, De\_par and DA in pre-frontal white matter, FA in CC-Genu, AWF and De\_per in CC-Body, Stroop and SDMT).

For the proposed approach, the raw representation is 286 dimensional, including 20 words for each of 9 MR metrics (AWF, DA, De\_par, De\_perp, FA, MD, AK, MK, RK) in Corpus Callosum, and 5 MR metrics in Thalamus (FA, MD, AK, MK, RK), and 6 demographic and neurocognitive features.
BoW approach achieved further improvement in accuracy to 91\%, and the optimum subset contains 9 features (which includes age, and (multiple) words listed in Table II.
\begin{table}[ht]
\centering
  \caption{List of Selected Features}
  \centering
\begin{tabular}{|m{6.2cm}|m{1.5cm}|}
\hline
Selected Features  & Classification Accuracy\\
\hline 
FA in Thalamus, MK in Thalamus, FA in Thalamus, Deperp in CC, MK in Thalamus, FA in CC, MD in CC, AWF in CC, FA in CC &   \ \ \ \ \ \ 91\% \\
\hline
\end{tabular}
\label{TblComp}
\end{table}

We have also evaluated the classification accuracy for feature subset of different size. 
Figure 3 shows the classification accuracies achieved by optimum subset of feature of dimension 1 to 9.
\begin{figure}[h]
\begin{center}
%\captionsetup{justification=centering}
    \includegraphics [scale=0.25] {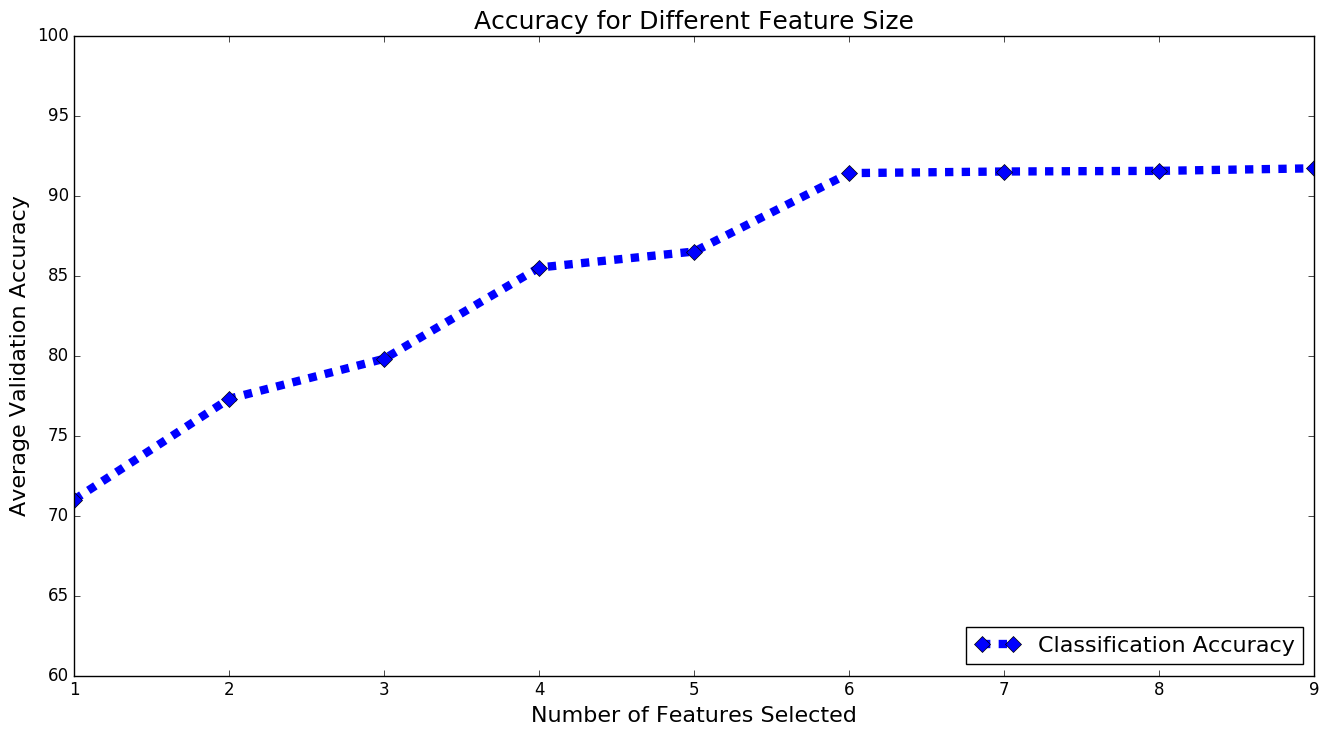}
\end{center}
\vspace{-0.2cm}
  \caption{Classification accuracy for feature set of different size.}
\end{figure}

Besides classification accuracy, we also report the sensitivity and specificity, which are important in the study of medical data analysis. 
The sensitivity and specificity are defined as in Eq. (1), where TP, FP, TN, and FN denote true positive, false positive, true negative, and false negative respectively. 
In our evaluation, we treat the mTBI subjects as positive. 
\begin{gather}
 \text{Sensitivity}= \frac{\text{TP}}{\text{TP+FN}} \ , 
\ \ \ \ \text{Specificity}= \frac{\text{TN}}{\text{TN+FP}} %%%% True Positive Rate 
\end{gather}
Figure 4 denotes the classification accuracies, sensitivities and specificities for different ratios of training samples (i.e. to keep different percentages of samples as training and the rest as test).
\begin{figure}[h]
\begin{center}
%\captionsetup{justification=centering}
    \includegraphics [scale=0.25] {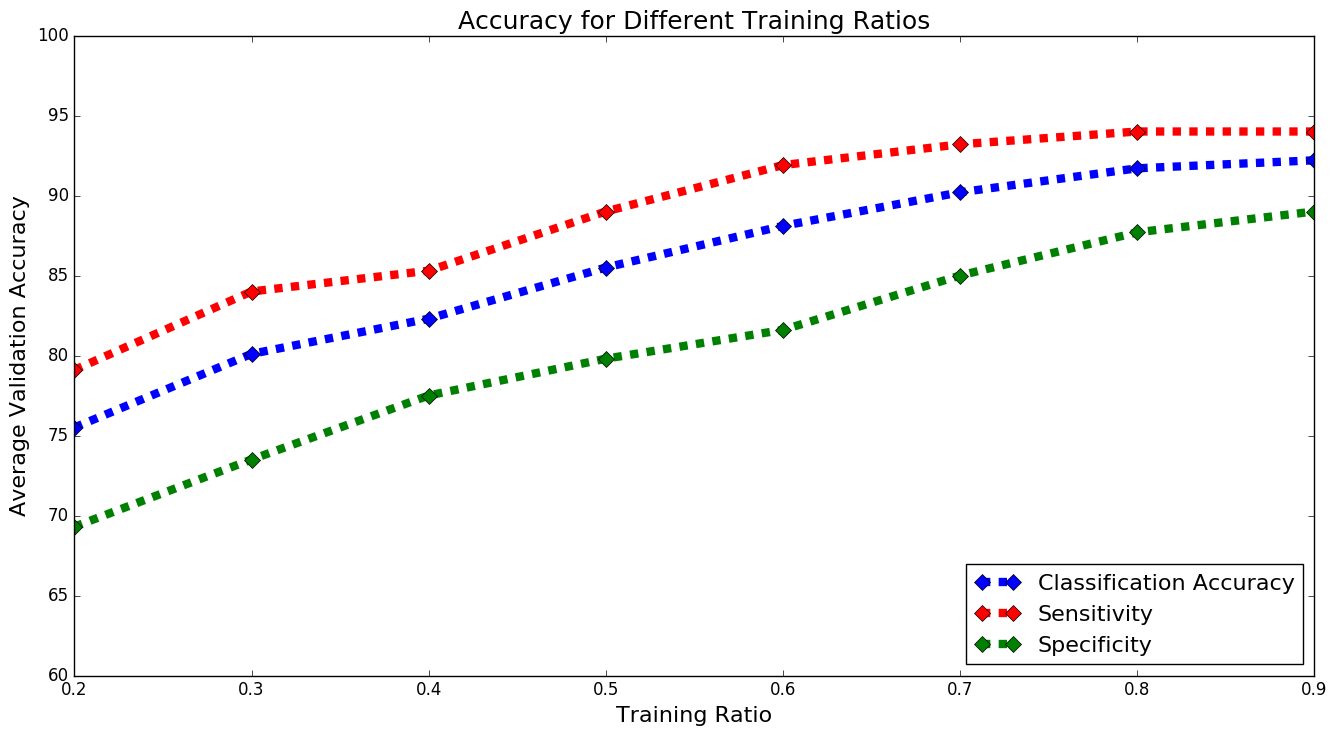}
\end{center}
\vspace{-0.2cm}
  \caption{Cross-validation  accuracy for different training ratios.}
\end{figure}

The classification accuracies using different approaches are  summarized in Table III.
\begin{table}[ht]
\centering
  \caption{Performance comparison for different approaches}
  \centering
\begin{tabular}{|m{5cm}|m{1.5cm}|}
\hline
Features used with SVM  & Classification Accuracy\\
\hline 
Single best feature &   \ \ \ \ \  72\% \\
\hline
The selected subset with 8 features   &  \ \ \ \  \ 80\%\\
\hline
BoW approach with 10 selected features   &  \ \ \ \ \ 91\%\\
\hline
\end{tabular}
\label{TblComp}
\end{table}

As we can see, BoW approach achieves significant improvement over mean value features. 
It is worth to mention that we also tried other classifiers such as neural network and logistic regression, but their performance was slightly worse than SVM, which is something expected on smaller datasets (neural network classification accuracy was around 86\% with the selected feature subset).
One possible explanation is that BoW features are able to find more discriminative pattern between mTBI and control cohorts. 
To verify this, we find the average histogram representation of patient subjects, and compare it with the average histogram representation of control subjects. 
These histograms and their difference are shown in Figure 5.
As we can see mTBI and control subjects have clear differences in some part of these representations.

\begin{figure}[h]
\begin{center}
%\captionsetup{justification=centering}
    \includegraphics [scale=0.21] {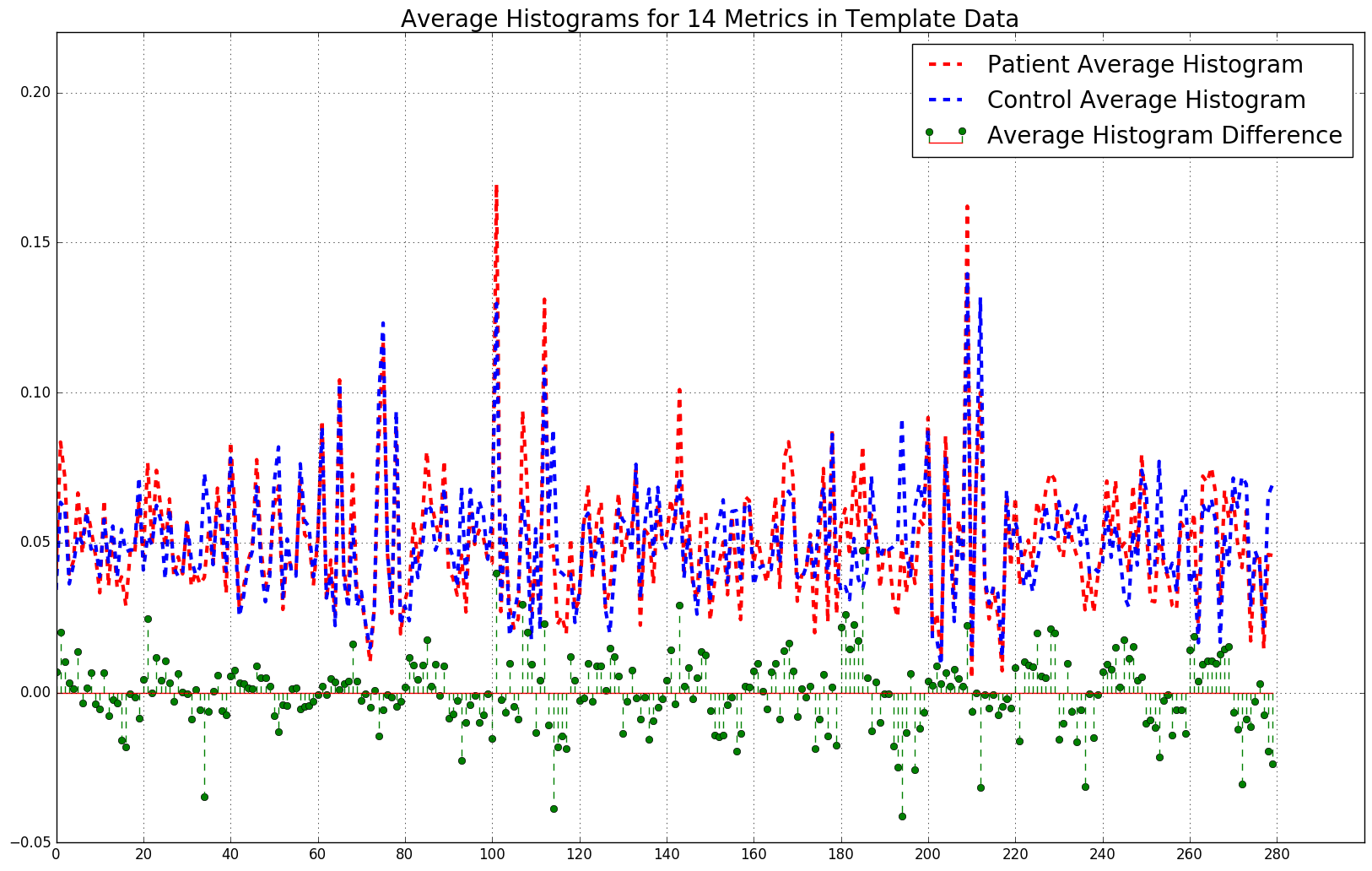}
\end{center}
\vspace{-0.2cm}
  \caption{The comparison between average histograms}
\end{figure}

We also provide the visualization of some of the visual words that are chosen by our algorithm in Figure 6. 
As we can see from this figure, these words correspond to different patterns in the brain.
\begin{figure}[h]
\begin{center}
%\captionsetup{justification=centering}
    \includegraphics [scale=0.37] {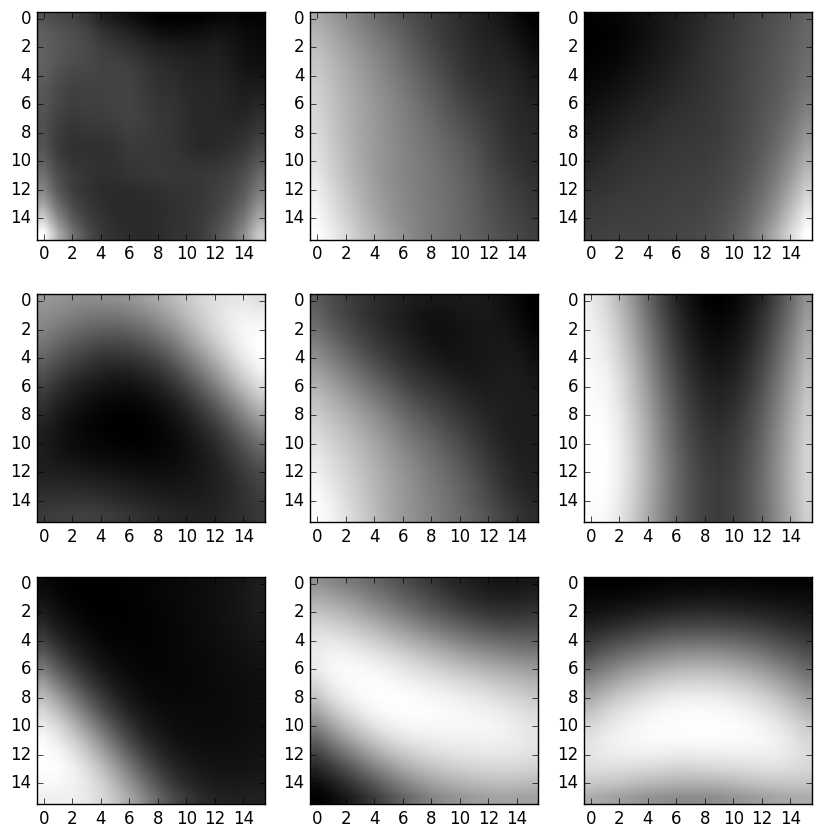}
\end{center}
\vspace{-0.2cm}
  \caption{The words chosen by the proposed algorithm}
\end{figure}

\section{Conclusions}
Here we show the application of bag of visual words on diffusion MR images for classification of patients with mTBI compared with controls.
In this approach, a set of visual features are learned from multiple MR metrics in two brain regions, and are used along two demographic features and four neurocognitive tests. 
Then greedy forward feature selection and support vector machine are used to perform classification.
We show that by learning visual features, we obtain significant gain over mean value features which were used previously.
%This shows the promise of feature learning for medical image classification.
These visual features can also be used for long-term outcome prediction of mTBI patients \cite{mtbi-pre}.

\section{Acknowledgement}
We would like to thank Cameron Johnson for his help on some part of this project.

% that's all folks
\end{document}